\documentclass{article}
\usepackage{arxiv}

\usepackage[utf8]{inputenc} 
\usepackage[T1]{fontenc}    
\usepackage{hyperref}       
\usepackage{url}            
\usepackage{booktabs}       
\usepackage{amsfonts}       
\usepackage{nicefrac}       
\usepackage{microtype}      
\usepackage{amsmath}
\usepackage{cleveref}       
\usepackage{lipsum}         
\usepackage{graphicx}
\usepackage{natbib}
\usepackage{doi}

\hypersetup{colorlinks,allcolors=blue}

\title{Activation-Based Active Learning for In-Context Learning: Challenges and Insights}

\date{}

\usepackage{authblk}

\author[1]{
	{Yaseen M.~Osman\thanks{\texttt{Correspondence: y.m.osman@soton.ac.uk}}}
}
\author[1]{
	{Geoff V.~Merrett}
}
\author[1]{
	{Stuart E.~Middleton}
}
\affil[1]{School of Electronics and Computer Science, University of Southampton, United Kingdom}

\hypersetup{
pdftitle={Activation-Based Active Learning for In-Context Learning: Challenges and Insights},
pdfsubject={cs.CL, cs.LG},
pdfauthor={Yaseen M.~Osman, Geoff V.~Merrett, Stuart E.~Middleton},
pdfkeywords={In-Context Learning, MLP Activations, Active Learning, Transformers, LLMs},
}

\begin{document}

\maketitle

\begin{abstract}
	Deep active learning has previously been explored for LLM in-context sample selection, but not with methods that utilise recent advances in understanding of transformer activations. In this paper, we test the hypothesis that model activations could provide a fine-grained signal to optimise the selection of in-context examples. We present a comprehensive analysis of MLP activation-based deep active learning methods applied to in-context learning, including how different attention masking strategies impact active learning across diverse classification and generative datasets, using both Llama-3.2-3B and Qwen2.5-3B base models. However, we find a negative result: MLP and embedding layer outputs, viewed through the lenses of massive activations or the first four moments, do not correlate with example quality or task performance. Specifically, the absolute Spearman correlation coefficient is at most 0.33 for all tasks and models we tested, showing that such activation-based sampling should not be used for in-context learning. We hypothesise that this may be due to superposition, whereby models represent more features than they have dimensionality, suggesting that methods like Sparse Autoencoders (SAEs) may be a promising future direction.\footnote{Our code and a link to our project page are available at \url{https://github.com/yaseenosman/activation-based-al}.}
\end{abstract}

\section{Introduction}
In-context learning is a popular alternative to fine-tuning for improving the performance of Large Language Models (LLMs). It is the process of providing examples to the LLM in the prompt rather than updating the model weights \citep{Brown2020}. Recent studies have looked at the impact and sensitivity of LLMs towards the provided in-context examples \citep{schoch-ji-2025-monte, li-qiu-2023-finding, pmlr-v202-ye23c, qin-etal-2024-context, rubin-etal-2022-learning, zhang-etal-2022-active, yao-etal-2024-samples, wu-etal-2023-self}.

Active learning \citep{Settles2009ActiveSurvey, Cohn1994, ein-dor-etal-2020-active} aims to optimise the data selection phase by selecting data that is likely to yield better model training performance. Active learning has been extensively studied for machine learning models, such as Support Vector Machines (SVMs) \citep{Tong2001ActiveApplications} and neural networks \citep{Cohn1994}. Recently, deep active learning has been used with deep learning models such as transformers \citep{Liu2022, Zhang2020, Shelmanov2019, ein-dor-etal-2020-active, bayer-etal-2026-activellm}.

Deep active learning has been also adapted for sampling in-context learning examples for LLMs through a variety of techniques, such as using model uncertainty \citep{margatina-etal-2023-active, wang2025monocle}, data diversity \citep{margatina-etal-2023-active, li-qiu-2023-finding, lee2025contrastive, Malik2025}, mixtures of both diversity and uncertainty \citep{mavromatis2023examples, treerath2026active}, similarity between in-context examples and test samples encoded through sentence encoders \citep{liu-etal-2022-makes, margatina-etal-2023-active}, training a specialised retrieval-based model \citep{luo2023dricl}, Monte Carlo sampling \citep{schoch-ji-2025-monte} and reinforcement learning \citep{zhang-etal-2022-active}.

The hypothesis of this paper is that model activations could provide a fine-grained signal to optimise the selection of in-context examples, grounded in both the transformer layer weights, which encode LLM pre-trained knowledge, and the relevance of the in-context example to the test sample. In this paper, we focus on MLP layer activations given the crucial role of the MLP layer in knowledge memorisation \citep{dong2025random, bricken2023monosemanticity, nanda2023fact}. We also explore adjusting causal masking \citep{behnamghader2024llmvec, lin-etal-2025-look} to better capture complex relationships and enable greater information flow between the candidate in-context learning example and test sample. 

The contribution in this paper is a comprehensive analysis of MLP activation-based active learning methods applied to in-context learning to date, covering a diverse range of datasets comprising both classification and generative tasks. We include a study examining how different LLM masking strategies impact active learning, including tailored masking methods for in-context learning sampling. Our results show that activation-based methods for active learning do not correlate directly with task performance. This is a negative result for our hypothesis. However, this study provides a contribution in terms of useful insights and challenges that we derive from our results to guide future research in this area.

\section{Related Work}
\paragraph{Attention Masking Adjustments}
Recent work explored removing the causal masking from pre-trained decoder-only transformer models to enhance their performance in encoder-based and embedding tasks by allowing the model to process full sequences bidirectionally \citep{behnamghader2024llmvec, lin-etal-2025-look}. Additionally, \citet{springer2025repetition} and \citet{leviathan2025prompt} achieved a similar effect by instead duplicating the prompt twice in the input. Other work focusing on studying and adjusting attention masking includes \citet{Barbero2024, barbero2025why, Swietojanski2023, Wu2024}.

\paragraph{In-Context Data Sampling}
Adjacently, for in-context learning examples sampling, \citet{liu-etal-2022-makes} and \citet{margatina-etal-2023-active} have utilised another model's embeddings to encode both examples and tests, then employed the k-nearest neighbours algorithm. Furthermore, sampling using the attention weights of the queried transformer model has been explored in the context of Retrieval-Augmented Generation (RAG) \citep{Chen2025} and Visual Instruction Tuning \citep{hu2025delta}. We consider these sampling techniques to be the most relevant.

\paragraph{Mechanistic Interpretability of Activations and In-Context Learning}
Several works have used Sparse Autoencoders (SAEs) to facilitate the interpretation of activations and mitigate difficulties from superposition \citep{Sharkey2022, bricken2023monosemanticity, huben2024sparse}. Superposition is when the model represents more features than its dimensionality, resulting in highly uninterpretable and entangled activation matrices \citep{elhage2022superposition}.

Notably, \citet{cho-hockenmaier-2025-toward} used SAEs to perform downstream steering to increase the in-context learning performance of a model. On the other hand, \citet{cho-etal-2025-versatility} utilised SAEs beyond interpretability and found that SAEs reconstruction loss correlates with prompt task performance for in-context learning settings.

Moreover, \citet{olsson2022context} attributed the emergence of the in-context learning ability to specialised attention heads called induction heads.

\section{Method}
In this section, we present our methodology for scoring in-context examples using active learning and how different types of masking impact sampling performance.

\subsection{Masking}
To allow for different information-flow variants between the example and the test sample, we explored various masks. Given a 1-shot prompt, these are:
\begin{itemize}
    \item \textbf{Causal Mask}, which blocks the attention from future tokens (i.e., default masking).
    \item \textbf{No Mask}, which removes the causal masking following \citep{lin-etal-2025-look}.
    \item \textbf{Row Test-Based}, which only masks the test \textbf{\textit{queries}} (i.e., rows of test tokens in attention weights).
    \item \textbf{Col Test-Based}, which only masks the test \textbf{\textit{keys}} (i.e., columns of test tokens in attention weights).
\end{itemize}

\begin{figure}[t]
  \includegraphics[width=\linewidth]{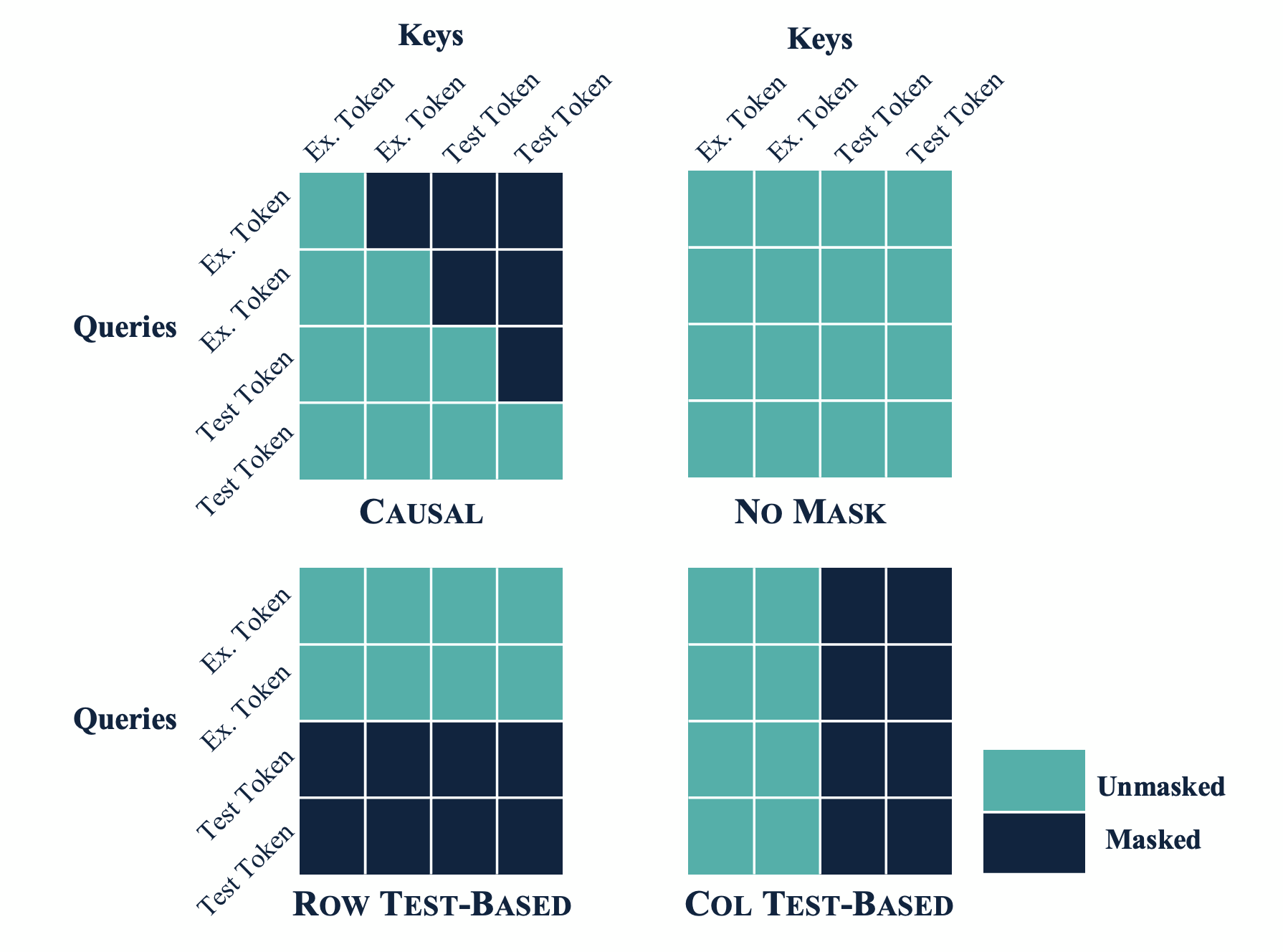}
  \caption{Shows a visualisation of the different attention maskings used in this paper using a prompt consisting of both example tokens (i.e., Ex. Token) and test tokens.}
  \label{fig:masks}
\end{figure}

Figure~\ref{fig:masks} depicts these maskings.

\subsection{In-Context Examples Scoring}
Once the masking of a model, containing $L$ layers, has been adjusted, given a set of candidate examples $M$ and test samples $N$, for each example $m \in M$, we construct $len(N)$ 1-shot prompts. Specifically, for an $n \in N$, $p_{m,n}$ is the concatenation of example $m$, a delimiter "\textbackslash n\textbackslash n", $n$. We then input this as a prompt to the model for a single forward pass, resulting in $L+1$ output hidden states as activation matrices (i.e., one for each MLP layer and for the embedding layer) $\{X_0, \dots, X_L\}$. The score of the prompt is given by:

\begin{equation}
    pt\_score(p_{m, n}) = \frac{1}{L+1} \sum_{l=0}^{L}metric(X_l),
\label{Eq:PromptScore}
\end{equation}

such that $metric()$ is a metric of choice to reduce an activation matrix to a score. In Appendix \ref{section:embeddinglayer}, we explore the removal of $X_0$ (i.e., the activations of the embedding layer). Additionally, we investigate the impact of each layer independently in Appendix \ref{section:perlayer}.

To get the overall score of $m$, we take the mean of all prompt scores that consist of $m$, as follows:

\begin{align}
    st\_score(m) = \frac{1}{len(N)} \sum_{n \in N} pt\_score(p_{m, n}).
    \label{Eq:ActivationScore}
\end{align}

\begin{figure}[t]
  \includegraphics[width=\linewidth]{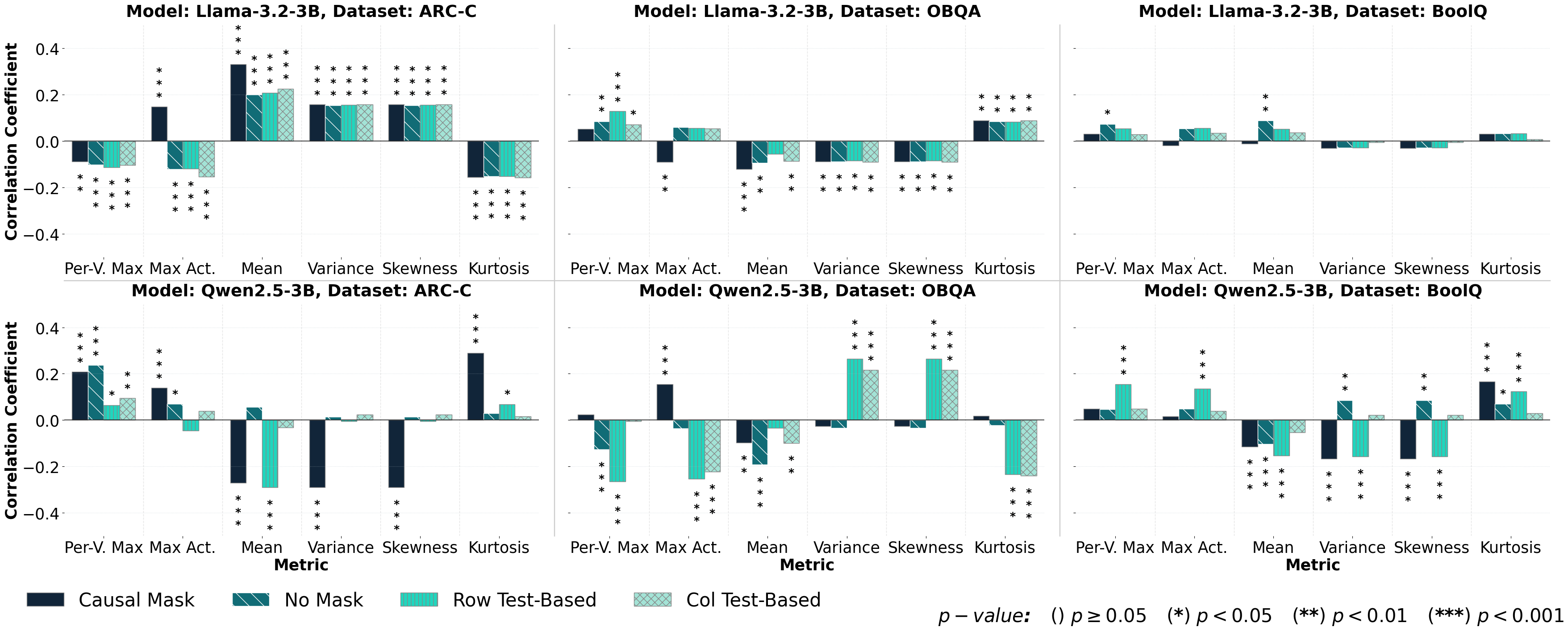}
  \caption {Shows the Spearman correlation and the associated $p$-values between task performance and the different activation-based metrics discussed. Results are for both Llama-3.2-3B and Qwen2.5-3B across three different classification-based datasets.}
    \label{fig:classification}
\end{figure}

\subsection{Metric Choices}
The choices of metrics used in Equation~\ref{Eq:PromptScore}, given $X_l$, are the first four moments: mean, variance, skewness, and kurtosis. These moments allow the capturing of various activation patterns. For the mean, we take the absolute value of $X_l$ before calculating its mean. This is because we empirically observe that the mean of MLP activations is close to zero, owing to the presence of many positive and negative activation values. We do not make any changes to the other three moments.

We use the largest massive activation, given its importance in model performance \citep{sun2024massive}, as an additional metric by selecting the activation with the largest absolute magnitude in $X_l$. We refer to this as the max activation. For completeness, we also explore the max activation, but for each feature vector and take the mean across all vectors in $X_l$. We refer to it as per-vector max.

In this section, we outline our experimental setup in detail before presenting our results. We conduct all our evaluations using the LM evaluation harness framework \citep{eval-harness}.

\subsection{Models}
We mainly experimented with two models, which are Llama-3.2-3B \citep{grattafiori2024llama}, and Qwen2.5-3B \citep{qwen2.5, qwen2}. These provide examples for two different families that are frequently used. We also explored Llama-3.1-8B and Llama-3.1-8B-Instruct.

\subsection{Datasets}
We use a diverse set of datasets, covering classification-based (i.e., two multiple-choice tasks and a true/false task) and generative-based tasks (i.e., a math reasoning task). These are ARC-Challenge (ARC-C) \citep{clark2018think}, OpenBookQA (OBQA) \citep{mihaylov-etal-2018-suit}, BoolQ \citep{clark-etal-2019-boolq, Wang2019SGlue}, and GSM8K \citep{cobbe2021training}, with the number of test samples, $len(N)$, being 1172, 500, 3270, and 1319, respectively. 

\subsection{Sampling and Evaluation}
For each dataset, we randomly sample a subset $M$ of 1000 in-context example candidates from its training split. However, for ARC-C, we use the full training split of 1119 samples as it is close to 1000.

Moreover, we calculate the $st\_score$ for all examples in $M$. We also evaluate each of the resulting $p_{m, n}$ 1-shot prompts on the task. Specifically, we evaluate task performance using exact match with strict match filtering for GSM8K and using accuracy for the other datasets. Finally, we compute the Spearman correlation coefficient between the task performance and $st\_score$ through utilising the SciPy implementation (i.e., $scipy.stats.spearmanr$) \citep{2020SciPy-NMeth}.

We focus solely on 1-shot in-context learning to simplify the exploration and reduce the computational cost of our experiment, thereby enabling a finer-grained analysis. However, our scoring method can be applied to few-shot settings. Lastly, we use a batch size of 1 for all our experiments on an Nvidia L4, except for GSM8K, which we run on an Nvidia L40 with a batch size of 64 due to the computational time of the generative task.

\section{Results}
\begin{figure}[t]
  \includegraphics[width=\linewidth]{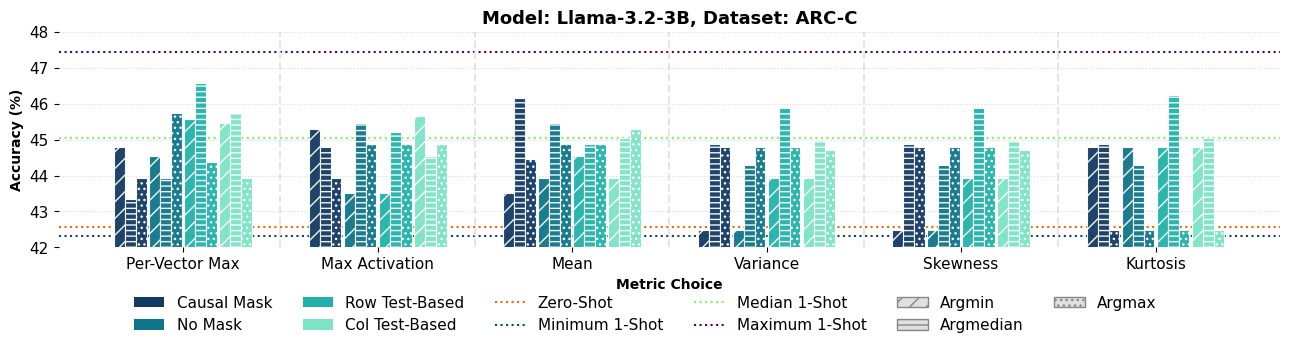}
  \caption{Shows the accuracy of Llama-3.2-3B on ARC-C when sampling examples with argmin, argmedian, and argmax metric values (column pattern) using the different masks (column colour). The accuracy does not align with argmin, argmedian, and argmax. The performances of the examples of the maximum, median and minimum accuracies are also included.}
  \label{fig:arccperformance}
\end{figure}

\begin{figure}[t]
  \includegraphics[width=\linewidth]{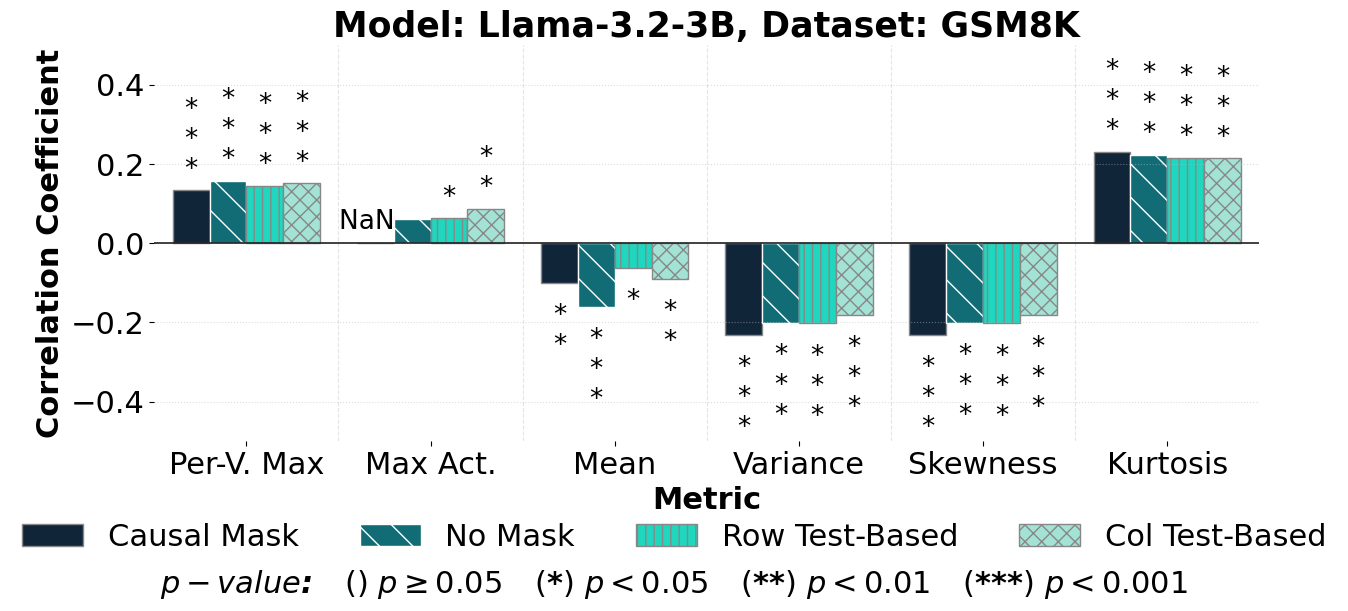}
  \caption{Shows the Spearman correlation and $p$-values between task performance and different activation-based metrics for GSM8K using Llama-3.2-3B.}
  \label{fig:gsm8k}
\end{figure}

In Figure~\ref{fig:classification}, we present the correlation coefficient across three classification-based datasets. For both Qwen2.5-3B and Llama-3.2-3B, a consistently low correlation coefficient is observed. Additionally, low $p$-values (i.e., $p < 0.001$) are observed across several experiments, reflecting a lack of correlation with high statistical significance. However, some results, especially when using Qwen2.5-3B or both models on BoolQ, have relatively higher $p$-values (i.e., $p \geq 0.05$), indicating lower statistical significance. We argue that the lack of correlation is persistent due to the very low correlation values in these results, which approach zero (i.e., $< 0.1$). Furthermore, the highest overall correlation value achieved is only around 0.33.

Figure~\ref{fig:arccperformance} includes the performance of examples that achieve the maximum, median, and minimum scores under each metric and mask combination on ARC-C using Llama-3.2-B, referred to as argmax, argmedian, and argmin, respectively. The true best example is never selected. Moreover, across all metrics and masks, performance and score magnitude do not align. Additionally, argmax and argmin values indicate high variability in 1-shot performance, and some examples drop performance even below the zero-shot accuracy, aligning with previous work, such as \citet{schoch-ji-2025-monte}.

\begin{figure}[t]
  \includegraphics[width=\linewidth]{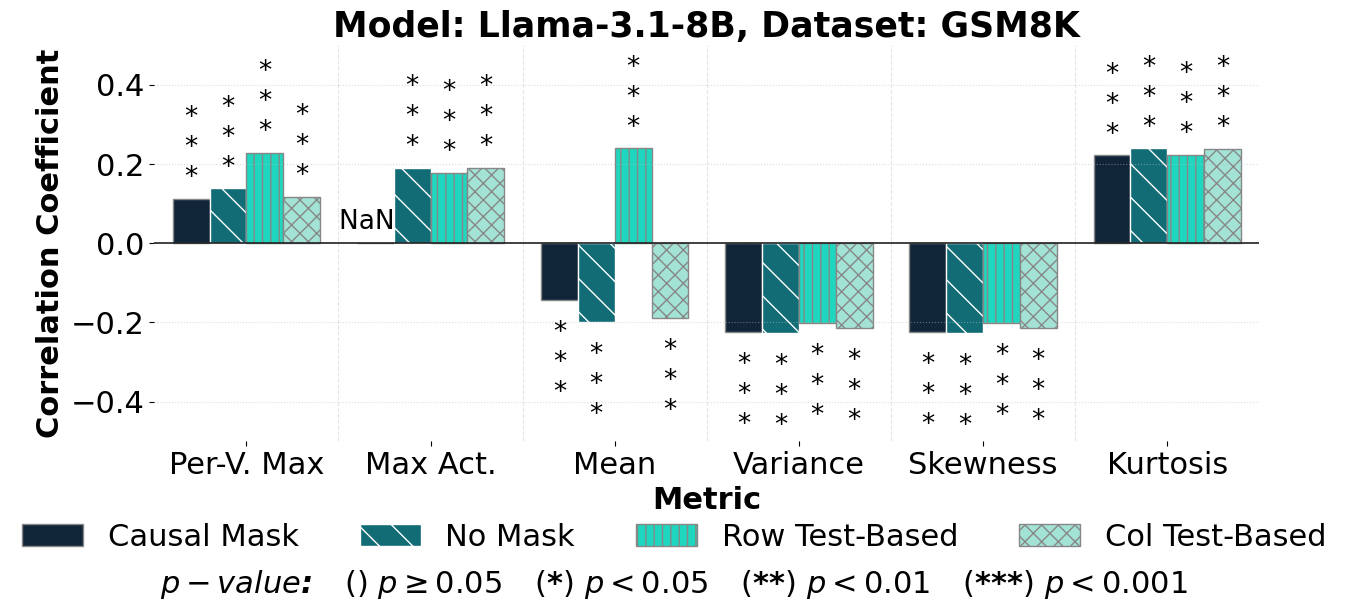} \vfill
  \includegraphics[width=\linewidth]{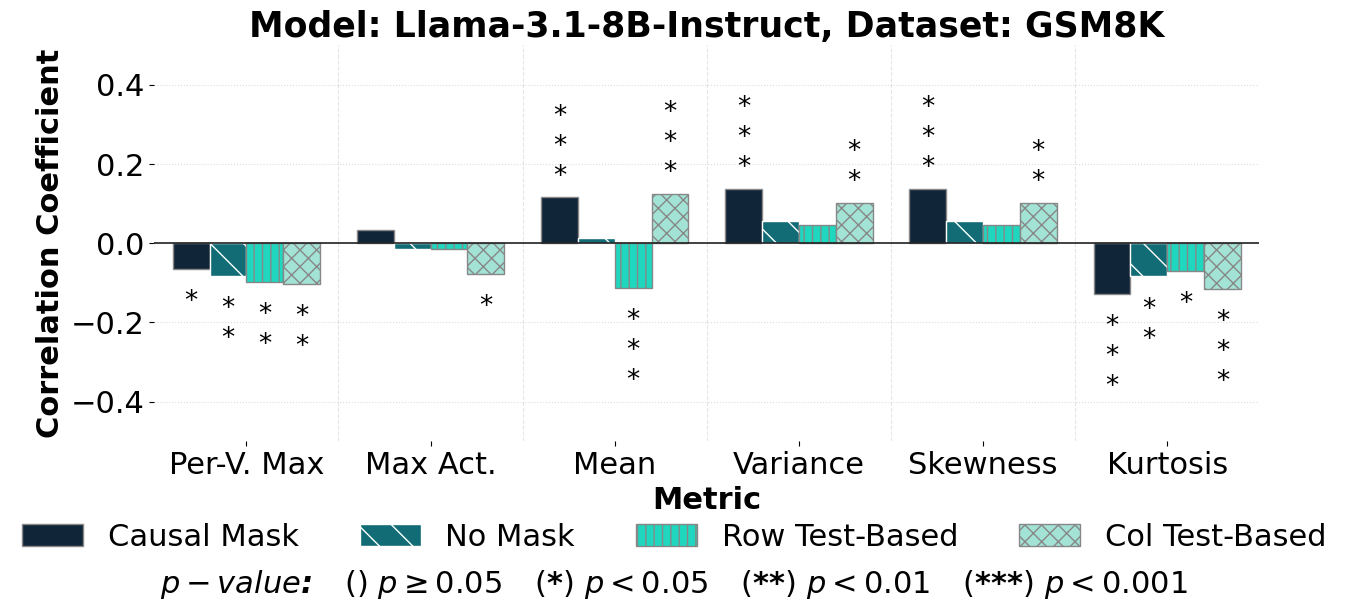}
  \caption {Shows the Spearman correlation and $p$-values between task performance and different activation-based metrics for GSM8K using Llama-3.1-8B and Llama-3.1-8B-Instruct.}
  \label{fig:8BInstruct}
\end{figure}

We expanded our Llama-3.2-3B to GSM8K example reasoning-based generative tasks in Figure~\ref{fig:gsm8k}. However, the lack of correlation persists across all masks. The NaN value resulting from using causal masking with the max activation metric is due to the largest massive activation value not changing across all prompts. This results in zero variance, which in turn results in an undefined Spearman correlation.

Additionally, we run our GSM8K experiments on both Llama-3.1-8B and Llama-3.1-8B-Instruct to cover a wider variety of model sizes and variants. The observation of no correlation is still maintained, as shown in Figure~\ref{fig:8BInstruct}. The absence of the NaN value in the case of the instruct variant (i.e., Llama-3.1-8B-Instruct) when computing the max activation metric for the causal mask is due to extremely low variance (i.e., around 7e-9) in the example scores. This is in contrast to the zero variance in the cases of Llama-3.1-8B and Llama-3.2-3B. Furthermore, we conduct auxiliary experiments in Appendix \ref{section:embeddinglayer} by excluding the activations of the embedding layer when conducting the $pt\_score$ computations. We also compute the scores and calculate the correlation for each layer of Llama-3.2-3B for ARC-C independently, which can be found in Appendix \ref{section:perlayer}.

Based on the above, we conclude that there is no reliable correlation between activation patterns and in-context example performance across the metrics we have explored. We hypothesise that this is due to the superposition phenomenon, resulting in each feature dimension representing complex, interconnected features between dimensions, requiring a non-linear filtering approach \citep{elhage2022superposition}. Hence, we highlight the use of Sparse Autoencoders (SAEs) \citep{huben2024sparse, cho-etal-2025-versatility} being an alternative to the metrics used in this paper, as a promising future direction to advance further.

\section{Conclusion}
In this work, we showed the significantly low correlation between task performance and in-context example scores using MLP and embedding layer activations (i.e., per-layer output hidden states) across a diverse set of tasks. Thus, we argue that MLP activations should not be used to sample in-context examples without further disentangling the MLP, and that practitioners should use other sampling methods that yield more reliable scores (e.g., model uncertainty or attention weights). We hope these findings contribute to both LLMs' explainability and active learning efforts.

\section*{Limitations}
Due to computational costs and to provide finer-grained analysis, we examine only 1-shot prompts in this paper. We expect results to hold for higher shot counts, but leave it for future work to confirm this.

\section*{Acknowledgments}
This work was supported by the UK Research and Innovation Centre for Doctoral Training in Machine Intelligence for Nano-electronic Devices and Systems [EP/S024298/1]. The authors acknowledge the use of the IRIDIS High Performance Computing Facility, and associated support services at the University of Southampton, in the completion of this work.

\bibliographystyle{unsrtnat}
\bibliography{references}

@inproceedings{behnamghader2024llmvec,
    title={{LLM}2Vec: Large Language Models Are Secretly Powerful Text Encoders},
    author={Parishad BehnamGhader and Vaibhav Adlakha and Marius Mosbach and Dzmitry Bahdanau and Nicolas Chapados and Siva Reddy},
    booktitle={First Conference on Language Modeling},
    year={2024},
    url={https://openreview.net/forum?id=IW1PR7vEBf}
}

@inproceedings{Brown2020,
    author = {Brown, Tom and Mann, Benjamin and Ryder, Nick and Subbiah, Melanie and Kaplan, Jared D and Dhariwal, Prafulla and Neelakantan, Arvind and Shyam, Pranav and Sastry, Girish and Askell, Amanda and Agarwal, Sandhini and Herbert-Voss, Ariel and Krueger, Gretchen and Henighan, Tom and Child, Rewon and Ramesh, Aditya and Ziegler, Daniel and Wu, Jeffrey and Winter, Clemens and Hesse, Chris and Chen, Mark and Sigler, Eric and Litwin, Mateusz and Gray, Scott and Chess, Benjamin and Clark, Jack and Berner, Christopher and McCandlish, Sam and Radford, Alec and Sutskever, Ilya and Amodei, Dario},
    booktitle = {Advances in Neural Information Processing Systems},
    editor = {H. Larochelle and M. Ranzato and R. Hadsell and M.F. Balcan and H. Lin},
    pages = {1877--1901},
    publisher = {Curran Associates, Inc.},
    title = {Language Models are Few-Shot Learners},
    url = {https://proceedings.neurips.cc/paper_files/paper/2020/file/1457c0d6bfcb4967418bfb8ac142f64a-Paper.pdf},
    volume = {33},
    year = {2020}
}

@inproceedings{Chen2025,
    author = {Chen, Shijie and Jimenez Gutierrez, Bernal and Su, Yu},
    booktitle = {International Conference on Learning Representations},
    editor = {Y. Yue and A. Garg and N. Peng and F. Sha and R. Yu},
    pages = {73047--73069},
    title = {Attention in Large Language Models Yields Efficient Zero-Shot Re-Rankers},
    url = {https://proceedings.iclr.cc/paper_files/paper/2025/file/b5b1890a7c1a79fe9b32b0f442347089-Paper-Conference.pdf},
    volume = {2025},
    year = {2025}
}

@inproceedings{sun2024massive,
    title={Massive Activations in Large Language Models},
    author={Mingjie Sun and Xinlei Chen and J Zico Kolter and Zhuang Liu},
    booktitle={First Conference on Language Modeling},
    year={2024},
    url={https://openreview.net/forum?id=F7aAhfitX6}
}

@inproceedings{huben2024sparse,
    title={Sparse Autoencoders Find Highly Interpretable Features in Language Models},
    author={Robert Huben and Hoagy Cunningham and Logan Riggs Smith and Aidan Ewart and Lee Sharkey},
    booktitle={The Twelfth International Conference on Learning Representations},
    year={2024},
    url={https://openreview.net/forum?id=F76bwRSLeK}
}

@article{elhage2022superposition,
    title={Toy Models of Superposition},
    author={Elhage, Nelson and Hume, Tristan and Olsson, Catherine and Schiefer, Nicholas and Henighan, Tom and Kravec, Shauna and Hatfield-Dodds, Zac and Lasenby, Robert and Drain, Dawn and Chen, Carol and Grosse, Roger and McCandlish, Sam and Kaplan, Jared and Amodei, Dario and Wattenberg, Martin and Olah, Christopher},
    year={2022},
    journal={Transformer Circuits Thread},
    url={https://transformer-circuits.pub/2022/toy_model/index.html}
}

@misc{eval-harness,
    author       = {Gao, Leo and Tow, Jonathan and Abbasi, Baber and Biderman, Stella and Black, Sid and DiPofi, Anthony and Foster, Charles and Golding, Laurence and Hsu, Jeffrey and Le Noac'h, Alain and Li, Haonan and McDonell, Kyle and Muennighoff, Niklas and Ociepa, Chris and Phang, Jason and Reynolds, Laria and Schoelkopf, Hailey and Skowron, Aviya and Sutawika, Lintang and Tang, Eric and Thite, Anish and Wang, Ben and Wang, Kevin and Zou, Andy},
    title        = {A framework for few-shot language model evaluation},
    month        = 12,
    year         = 2023,
    publisher    = {Zenodo},
    version      = {v0.4.0},
    doi          = {10.5281/zenodo.10256836},
    url          = {https://zenodo.org/records/10256836}
}

@inproceedings{Wang2019SGlue,
    author = {Wang, Alex and Pruksachatkun, Yada and Nangia, Nikita and Singh, Amanpreet and Michael, Julian and Hill, Felix and Levy, Omer and Bowman, Samuel},
    booktitle = {Advances in Neural Information Processing Systems},
    editor = {H. Wallach and H. Larochelle and A. Beygelzimer and F. d\textquotesingle Alch\'{e}-Buc and E. Fox and R. Garnett},
    pages = {},
    publisher = {Curran Associates, Inc.},
    title = {SuperGLUE: A Stickier Benchmark for General-Purpose Language Understanding Systems},
    url = {https://proceedings.neurips.cc/paper_files/paper/2019/file/4496bf24afe7fab6f046bf4923da8de6-Paper.pdf},
    volume = {32},
    year = {2019}
}

@article{clark2018think,
    title={Think you have solved question answering? try arc, the ai2 reasoning challenge},
    author={Clark, Peter and Cowhey, Isaac and Etzioni, Oren and Khot, Tushar and Sabharwal, Ashish and Schoenick, Carissa and Tafjord, Oyvind},
    journal={arXiv preprint arXiv:1803.05457},
    year={2018}
}

@article{cobbe2021training,
    title={Training verifiers to solve math word problems},
    author={Cobbe, Karl and Kosaraju, Vineet and Bavarian, Mohammad and Chen, Mark and Jun, Heewoo and Kaiser, Lukasz and Plappert, Matthias and Tworek, Jerry and Hilton, Jacob and Nakano, Reiichiro and others},
    journal={arXiv preprint arXiv:2110.14168},
    year={2021}
}

@inproceedings{springer2025repetition,
    title={Repetition Improves Language Model Embeddings},
    author={Jacob Mitchell Springer and Suhas Kotha and Daniel Fried and Graham Neubig and Aditi Raghunathan},
    booktitle={The Thirteenth International Conference on Learning Representations},
    year={2025},
    url={https://openreview.net/forum?id=Ahlrf2HGJR}
}

@article{leviathan2025prompt,
    title={Prompt Repetition Improves Non-Reasoning LLMs},
    author={Leviathan, Yaniv and Kalman, Matan and Matias, Yossi},
    journal={arXiv preprint arXiv:2512.14982},
    year={2025}
}

@article{wang2025monocle,
    title={Monocle: Hybrid Local-Global In-Context Evaluation for Long-Text Generation with Uncertainty-Based Active Learning},
    author={Wang, Xiaorong and Yang, Ting and Zhang, Zhu and Wang, Shuo and Zhou, Zihan and Yang, Liner and Liu, Zhiyuan and Sun, Maosong},
    journal={arXiv preprint arXiv:2505.20195},
    year={2025}
}

@inproceedings{luo2023dricl,
    title={Dr.{ICL}: Demonstration-Retrieved In-context Learning},
    author={Man Luo and Xin Xu and Zhuyun Dai and Panupong Pasupat and Mehran Kazemi and Chitta Baral and Vaiva Imbrasaite and Vincent Y Zhao},
    booktitle={R0-FoMo:Robustness of Few-shot and Zero-shot Learning in Large Foundation Models},
    year={2023},
    url={https://openreview.net/forum?id=NDNb6L5xjI}
}

@article{mavromatis2023examples,
  title={Which examples to annotate for in-context learning? towards effective and efficient selection},
  author={Mavromatis, Costas and Srinivasan, Balasubramaniam and Shen, Zhengyuan and Zhang, Jiani and Rangwala, Huzefa and Faloutsos, Christos and Karypis, George},
  journal={arXiv preprint arXiv:2310.20046},
  year={2023}
}

@article{treerath2026active,
  title={Active In-Context Learning for Tabular Foundation Models},
  author={Treerath, Wilailuck and Pittorino, Fabrizio},
  journal={arXiv preprint arXiv:2603.27385},
  year={2026}
}

@inproceedings{lee2025contrastive,
    title={Contrastive In-Context Learning with Active Memory for Task Planning},
    author={Jiho Lee and Kyungjae Lee and Eunwoo Kim},
    booktitle={Workshop on Making Sense of Data in Robotics: Composition, Curation, and Interpretability at Scale at CoRL 2025},
    year={2025},
    url={https://openreview.net/forum?id=NaayWmHgL1}
}

@inproceedings{Malik2025,
    author = {Malik, Vijit and Pande, Atul and Majumder, Anirban},
    title = {IclForge: Enhancing In-Context Learning with Evolutionary Algorithms under Budgeted Annotation},
    year = {2025},
    isbn = {9798400720406},
    publisher = {Association for Computing Machinery},
    address = {New York, NY, USA},
    url = {https://doi.org/10.1145/3746252.3761536},
    doi = {10.1145/3746252.3761536},
    booktitle = {Proceedings of the 34th ACM International Conference on Information and Knowledge Management},
    pages = {5931–5938},
    numpages = {8},
    location = {Seoul, Republic of Korea},
    series = {CIKM '25}
}

@article{Liu2022,
    author = {Mingyi Liu and Zhiying Tu and Tong Zhang and Tonghua Su and Xiaofei Xu and Zhongjie Wang},
    doi = {10.1007/s11063-021-10737-x},
    issn = {1370-4621},
    issue = {3},
    journal = {Neural Processing Letters},
    month = {6},
    pages = {2433-2454},
    title = {LTP: A New Active Learning Strategy for CRF-Based Named Entity Recognition},
    volume = {54},
    year = {2022}
}

@inproceedings{Zhang2020,
    author = {Zhang, Leihan and Zhang, Le},
    title = {An Ensemble Deep Active Learning Method for Intent Classification},
    year = {2020},
    isbn = {9781450376273},
    publisher = {Association for Computing Machinery},
    address = {New York, NY, USA},
    url = {https://doi.org/10.1145/3374587.3374611},
    doi = {10.1145/3374587.3374611},
    booktitle = {Proceedings of the 2019 3rd International Conference on Computer Science and Artificial Intelligence},
    pages = {107–111},
    numpages = {5},
    location = {Normal, IL, USA},
    series = {CSAI '19}
}

@INPROCEEDINGS{Shelmanov2019,
    author={Shelmanov, Artem and Liventsev, Vadim and Kireev, Danil and Khromov, Nikita and Panchenko, Alexander and Fedulova, Irina and Dylov, Dmitry V.},
    booktitle={2019 IEEE International Conference on Bioinformatics and Biomedicine (BIBM)}, 
    title={Active Learning with Deep Pre-trained Models for Sequence Tagging of Clinical and Biomedical Texts}, 
    year={2019},
    volume={},
    number={},
    pages={482-489},
    doi={10.1109/BIBM47256.2019.8983157},
    ISSN={},
    month={Nov},
}

@techreport{Settles2009ActiveSurvey,
    Author = {Burr Settles},
    Institution = {University of Wisconsin--Madison},
    Number = {1648},
    Title = {Active Learning Literature Survey},
    Type = {Computer Sciences Technical Report},
    Year = {2009},
}

@phdthesis{Tong2001ActiveApplications,
    title = {{Active learning: Theory and applications}},
    year = {2001},
    booktitle = {ProQuest Dissertations and Theses},
    author = {Tong, Simon},
    pages = {184},
    url = {https://www.proquest.com/dissertations-theses/active-learning-theory-applications/docview/304730539/se-2},
    isbn = {978-0-493-40474-5},
    language = {English},
    school={Stanford University}
}

@inproceedings{Cohn1994,
    author = {Cohn, David and Ghahramani, Zoubin and Jordan, Michael},
    booktitle = {Advances in Neural Information Processing Systems},
    editor = {G. Tesauro and D. Touretzky and T. Leen},
    pages = {},
    publisher = {MIT Press},
    title = {Active Learning with Statistical Models},
    url = {https://proceedings.neurips.cc/paper_files/paper/1994/file/7f975a56c761db6506eca0b37ce6ec87-Paper.pdf},
    volume = {7},
    year = {1994}
}

@InProceedings{pmlr-v202-ye23c,
    title = 	 {Compositional Exemplars for In-context Learning},
    author =       {Ye, Jiacheng and Wu, Zhiyong and Feng, Jiangtao and Yu, Tao and Kong, Lingpeng},
    booktitle = 	 {Proceedings of the 40th International Conference on Machine Learning},
    pages = 	 {39818--39833},
    year = 	 {2023},
    editor = 	 {Krause, Andreas and Brunskill, Emma and Cho, Kyunghyun and Engelhardt, Barbara and Sabato, Sivan and Scarlett, Jonathan},
    volume = 	 {202},
    series = 	 {Proceedings of Machine Learning Research},
    month = 	 {23--29 Jul},
    publisher =    {PMLR},
    url = 	 {https://proceedings.mlr.press/v202/ye23c.html}
}

@article{dong2025random,
    title={Is Random Attention Sufficient for Sequence Modeling? Disentangling Trainable Components in the Transformer},
    author={Dong, Yihe and Noci, Lorenzo and Khodak, Mikhail and Li, Mufan},
    journal={arXiv preprint arXiv:2506.01115},
    year={2025}
}

@article{bricken2023monosemanticity,
    title={Towards Monosemanticity: Decomposing Language Models With Dictionary Learning},
    author={Bricken, Trenton and Templeton, Adly and Batson, Joshua and Chen, Brian and Jermyn, Adam and Conerly, Tom and Turner, Nick and Anil, Cem and Denison, Carson and Askell, Amanda and Lasenby, Robert and Wu, Yifan and Kravec, Shauna and Schiefer, Nicholas and Maxwell, Tim and Joseph, Nicholas and Hatfield-Dodds, Zac and Tamkin, Alex and Nguyen, Karina and McLean, Brayden and Burke, Josiah E and Hume, Tristan and Carter, Shan and Henighan, Tom and Olah, Christopher},
    year={2023},
    journal={Transformer Circuits Thread},
    url={https://transformer-circuits.pub/2023/monosemantic-features/index.html}
}

@inproceedings{nanda2023fact,
    title={Fact finding: Attempting to reverse-engineer factual recall on the neuron level},
    author={Nanda, Neel and Rajamanoharan, Senthooran and Kram{\'a}r, J{\'a}nos and Shah, Rohin},
    booktitle={Alignment Forum},
    volume={6},
    year={2023}
}

@ARTICLE{2020SciPy-NMeth,
    author  = {Virtanen, Pauli and Gommers, Ralf and Oliphant, Travis E. and
            Haberland, Matt and Reddy, Tyler and Cournapeau, David and
            Burovski, Evgeni and Peterson, Pearu and Weckesser, Warren and
            Bright, Jonathan and {van der Walt}, St{\'e}fan J. and
            Brett, Matthew and Wilson, Joshua and Millman, K. Jarrod and
            Mayorov, Nikolay and Nelson, Andrew R. J. and Jones, Eric and
            Kern, Robert and Larson, Eric and Carey, C J and
            Polat, {\.I}lhan and Feng, Yu and Moore, Eric W. and
            {VanderPlas}, Jake and Laxalde, Denis and Perktold, Josef and
            Cimrman, Robert and Henriksen, Ian and Quintero, E. A. and
            Harris, Charles R. and Archibald, Anne M. and
            Ribeiro, Ant{\^o}nio H. and Pedregosa, Fabian and
            {van Mulbregt}, Paul and {SciPy 1.0 Contributors}},
    title   = {{{SciPy} 1.0: Fundamental Algorithms for Scientific
            Computing in Python}},
    journal = {Nature Methods},
    year    = {2020},
    volume  = {17},
    pages   = {261--272},
    adsurl  = {https://rdcu.be/b08Wh},
    doi     = {10.1038/s41592-019-0686-2},
}

@misc{qwen2.5,
    title = {Qwen2.5: A Party of Foundation Models},
    url = {https://qwenlm.github.io/blog/qwen2.5/},
    author = {Qwen Team},
    month = {September},
    year = {2024}
}

@article{qwen2,
    title={Qwen2 Technical Report}, 
    author={An Yang and Baosong Yang and Binyuan Hui and Bo Zheng and Bowen Yu and Chang Zhou and Chengpeng Li and Chengyuan Li and Dayiheng Liu and Fei Huang and Guanting Dong and Haoran Wei and Huan Lin and Jialong Tang and Jialin Wang and Jian Yang and Jianhong Tu and Jianwei Zhang and Jianxin Ma and Jin Xu and Jingren Zhou and Jinze Bai and Jinzheng He and Junyang Lin and Kai Dang and Keming Lu and Keqin Chen and Kexin Yang and Mei Li and Mingfeng Xue and Na Ni and Pei Zhang and Peng Wang and Ru Peng and Rui Men and Ruize Gao and Runji Lin and Shijie Wang and Shuai Bai and Sinan Tan and Tianhang Zhu and Tianhao Li and Tianyu Liu and Wenbin Ge and Xiaodong Deng and Xiaohuan Zhou and Xingzhang Ren and Xinyu Zhang and Xipin Wei and Xuancheng Ren and Yang Fan and Yang Yao and Yichang Zhang and Yu Wan and Yunfei Chu and Yuqiong Liu and Zeyu Cui and Zhenru Zhang and Zhihao Fan},
    journal={arXiv preprint arXiv:2407.10671},
    year={2024}
}

@article{grattafiori2024llama,
    title={The llama 3 herd of models},
    author={Grattafiori, Aaron and Dubey, Abhimanyu and Jauhri, Abhinav and Pandey, Abhinav and Kadian, Abhishek and Al-Dahle, Ahmad and Letman, Aiesha and Mathur, Akhil and Schelten, Alan and Vaughan, Alex and others},
    journal={arXiv preprint arXiv:2407.21783},
    year={2024}
}

@article{hu2025delta,
    title={$\Delta$-AttnMask: Attention-Guided Masked Hidden States for Efficient Data Selection and Augmentation},
    author={Hu, Jucheng and Yang, Suorong and Zhou, Dongzhan},
    journal={arXiv preprint arXiv:2508.09199},
    year={2025}
}

@inproceedings{Wu2024,
    author = {Wu, Xinyi and Ajorlou, Amir and Wang, Yifei and Jegelka, Stefanie and Jadbabaie, Ali},
    booktitle = {Advances in Neural Information Processing Systems},
    doi = {10.52202/079017-0472},
    editor = {A. Globerson and L. Mackey and D. Belgrave and A. Fan and U. Paquet and J. Tomczak and C. Zhang},
    pages = {14774--14809},
    publisher = {Curran Associates, Inc.},
    title = {On the Role of Attention Masks and LayerNorm in Transformers},
    url = {https://proceedings.neurips.cc/paper_files/paper/2024/file/1ac3030fc57850b0fb11dfe9d4880ad7-Paper-Conference.pdf},
    volume = {37},
    year = {2024}
}

@INPROCEEDINGS{Swietojanski2023,
    author={Swietojanski, Pawel and Braun, Stefan and Can, Dogan and Da Silva, Thiago Fraga and Ghoshal, Arnab and Hori, Takaaki and Hsiao, Roger and Mason, Henry and McDermott, Erik and Silovsky, Honza and Travadi, Ruchir and Zhuang, Xiaodan},
    booktitle={ICASSP 2023 - 2023 IEEE International Conference on Acoustics, Speech and Signal Processing (ICASSP)}, 
    title={Variable Attention Masking for Configurable Transformer Transducer Speech Recognition}, 
    year={2023},
    volume={},
    number={},
    pages={1-5},
    doi={10.1109/ICASSP49357.2023.10094588},
    ISSN={2379-190X},
    month={June},
}

@inproceedings{Barbero2024,
     author = {Barbero, Federico and Banino, Andrea and Kapturowski, Steven and Kumaran, Dharshan and Ara\'{u}jo, Jo\~{a}o G.M. and Vitvitskyi, Alex and Pascanu, Razvan and Veli\v{c}kovi\'{c}, Petar},
     booktitle = {Advances in Neural Information Processing Systems},
     doi = {10.52202/079017-3114},
     editor = {A. Globerson and L. Mackey and D. Belgrave and A. Fan and U. Paquet and J. Tomczak and C. Zhang},
     pages = {98111--98142},
     publisher = {Curran Associates, Inc.},
     title = {Transformers need glasses! Information over-squashing in language tasks},
     url = {https://proceedings.neurips.cc/paper_files/paper/2024/file/b1d35561c4a4a0e0b6012b2af531e149-Paper-Conference.pdf},
     volume = {37},
     year = {2024}
}

@inproceedings{barbero2025why,
    title={Why do {LLM}s attend to the first token?},
    author={Federico Barbero and Alvaro Arroyo and Xiangming Gu and Christos Perivolaropoulos and Petar Veli{\v{c}}kovi{\'c} and Razvan Pascanu and Michael M. Bronstein},
    booktitle={Second Conference on Language Modeling},
    year={2025},
    url={https://openreview.net/forum?id=tu4dFUsW5z}
}

@article{olsson2022context,
    title={In-context Learning and Induction Heads},
    author={Olsson, Catherine and Elhage, Nelson and Nanda, Neel and Joseph, Nicholas and DasSarma, Nova and Henighan, Tom and Mann, Ben and Askell, Amanda and Bai, Yuntao and Chen, Anna and Conerly, Tom and Drain, Dawn and Ganguli, Deep and Hatfield-Dodds, Zac and Hernandez, Danny and Johnston, Scott and Jones, Andy and Kernion, Jackson and Lovitt, Liane and Ndousse, Kamal and Amodei, Dario and Brown, Tom and Clark, Jack and Kaplan, Jared and McCandlish, Sam and Olah, Chris},
    year={2022},
    journal={Transformer Circuits Thread},
    note={https://transformer-circuits.pub/2022/in-context-learning-and-induction-heads/index.html}
}

@misc{Sharkey2022,
    author       = {Sharkey, Lee and Braun, Dan and Millidge, Beren},
    title        = {[Interim research report] Taking features out of superposition with sparse autoencoders},
    month        = 12,
    year         = 2022,
    publisher    = {AI Alignment Forum},
    url          = {https://www.lesswrong.com/posts/z6QQJbtpkEAX3Aojj/interim-research-report-taking-features-out-of-superposition}
}

@inproceedings{lin-etal-2025-look,title = "Look Both Ways and No Sink: Converting {LLM}s into Text Encoders without Training",author = "Lin, Ziyong and Wu, Haoyi and Wang, Shu and Tu, Kewei and Zheng, Zilong and Jia, Zixia",editor = "Che, Wanxiang and Nabende, Joyce and Shutova, Ekaterina and Pilehvar, Mohammad Taher",booktitle = "Proceedings of the 63rd Annual Meeting of the Association for Computational Linguistics (Volume 1: Long Papers)",month = jul,year = "2025",address = "Vienna, Austria",publisher = "Association for Computational Linguistics",url = "https://aclanthology.org/" # {2025.acl-long.1113/},doi = "10.18653/v1/2025.acl-long.1113",pages = "22839--22853",ISBN = "979-8-89176-251-0"}

@inproceedings{margatina-etal-2023-active,title = "Active Learning Principles for In-Context Learning with Large Language Models",author = "Margatina, Katerina and Schick, Timo and Aletras, Nikolaos and Dwivedi-Yu, Jane",editor = "Bouamor, Houda and Pino, Juan and Bali, Kalika",booktitle = "Findings of the Association for Computational Linguistics: EMNLP 2023",month = dec,year = "2023",address = "Singapore",publisher = "Association for Computational Linguistics",url = "https://aclanthology.org/" # {2023.findings-emnlp.334/},doi = "10.18653/v1/2023.findings-emnlp.334",pages = "5011--5034"}

@inproceedings{liu-etal-2022-makes,title = "What Makes Good In-Context Examples for {GPT}-3?",author = "Liu, Jiachang and Shen, Dinghan and Zhang, Yizhe and Dolan, Bill and Carin, Lawrence and Chen, Weizhu",editor = "Agirre, Eneko and Apidianaki, Marianna and Vuli{\'c}, Ivan",booktitle = "Proceedings of Deep Learning Inside Out (DeeLIO 2022): The 3rd Workshop on Knowledge Extraction and Integration for Deep Learning Architectures",month = may,year = "2022",address = "Dublin, Ireland and Online",publisher = "Association for Computational Linguistics",url = "https://aclanthology.org/" # {2022.deelio-1.10/},doi = "10.18653/v1/2022.deelio-1.10",pages = "100--114"}

@inproceedings{mihaylov-etal-2018-suit,title = "Can a Suit of Armor Conduct Electricity? A New Dataset for Open Book Question Answering",author = "Mihaylov, Todor and Clark, Peter and Khot, Tushar and Sabharwal, Ashish",editor = "Riloff, Ellen and Chiang, David and Hockenmaier, Julia and Tsujii, Jun{'}ichi",booktitle = "Proceedings of the 2018 Conference on Empirical Methods in Natural Language Processing",month = oct # "-" # nov,year = "2018",address = "Brussels, Belgium",publisher = "Association for Computational Linguistics",url = "https://aclanthology.org/" # {D18-1260/},doi = "10.18653/v1/D18-1260",pages = "2381--2391"}

\appendix
\section{Exclusion of the Embedding Layer} \label{section:embeddinglayer}
\begin{figure}[t]
  \includegraphics[width=\linewidth]{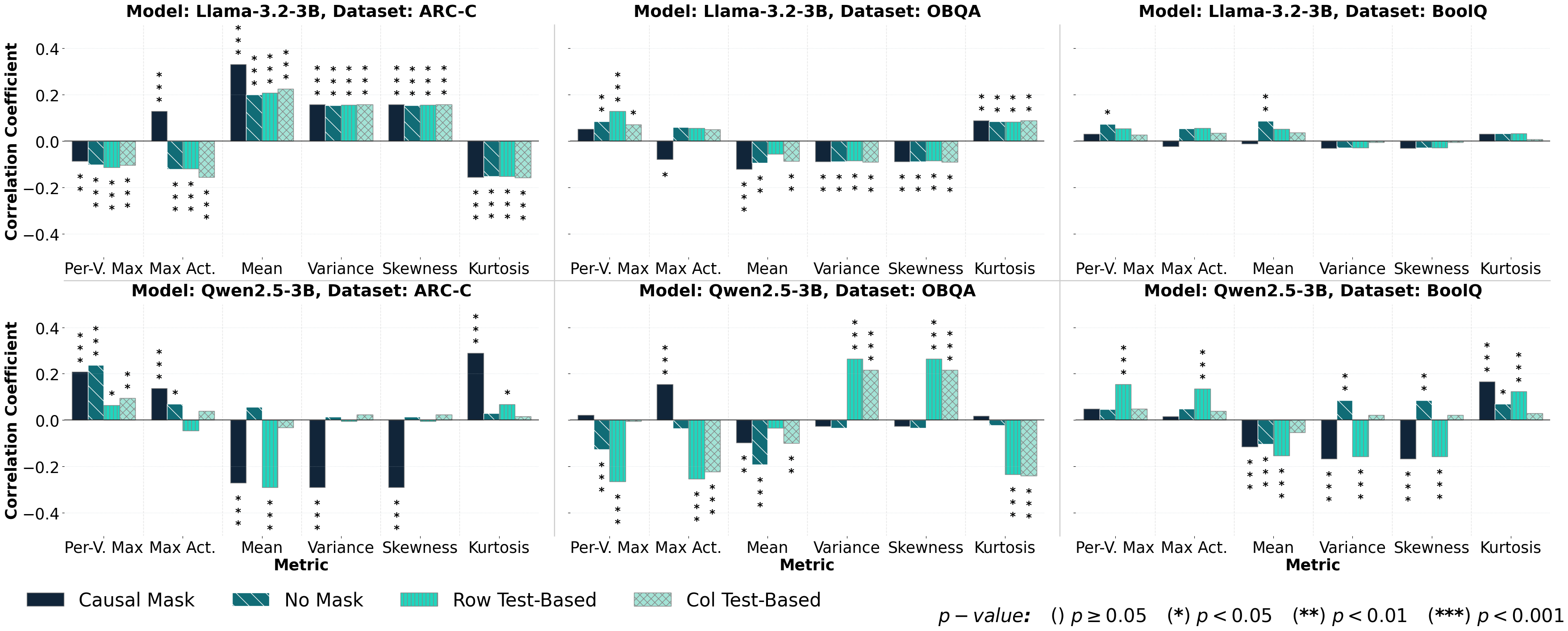}
  \caption {Shows the Spearman correlation and the associated $p$-values between task performance and the different activation-based metrics discussed. We remove the embedding layer from the $pt\_score$ computation. Results are for both Llama-3.2-3B and Qwen2.5-3B across three different classification-based datasets.}
    \label{fig:noembedding_classification}
\end{figure}

\begin{figure}[t]
  \includegraphics[width=\linewidth]{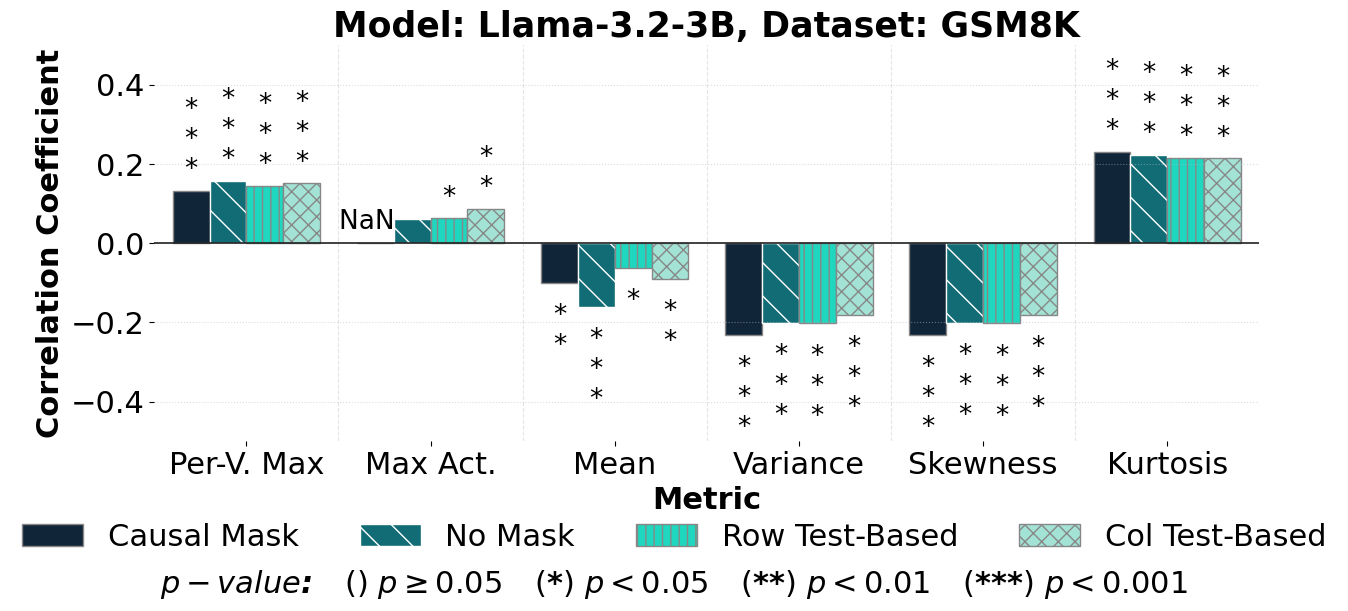}
  \caption{Shows the Spearman correlation and $p$-values between task performance and different activation-based metrics for GSM8K. We remove the embedding layer from the $pt\_score$ computation. Results are for Llama-3.2-3B.}
  \label{fig:noembedding_gsm8k}
\end{figure}

We present here the results of excluding the activations of the embedding layer (i.e., $X_0$) from Equation \ref{Eq:PromptScore}. For both Llama-3.2-3B and Qwen2.5-3B, we find that the lack of correlation is still maintained as shown in Figures \ref{fig:noembedding_classification} and \ref{fig:noembedding_gsm8k}.

\section{Per-Layer Analysis} \label{section:perlayer}
\begin{figure}[t]
  \includegraphics[width=\linewidth]{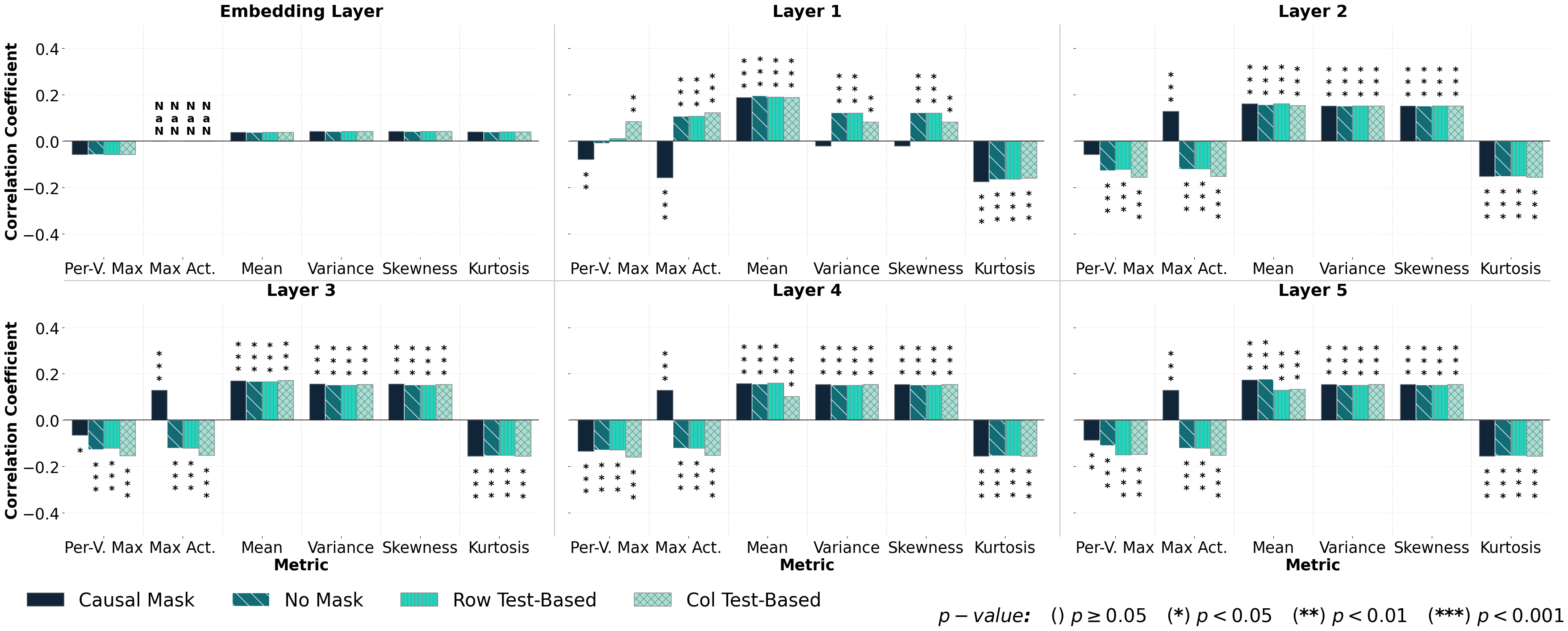}
  \includegraphics[width=\linewidth]{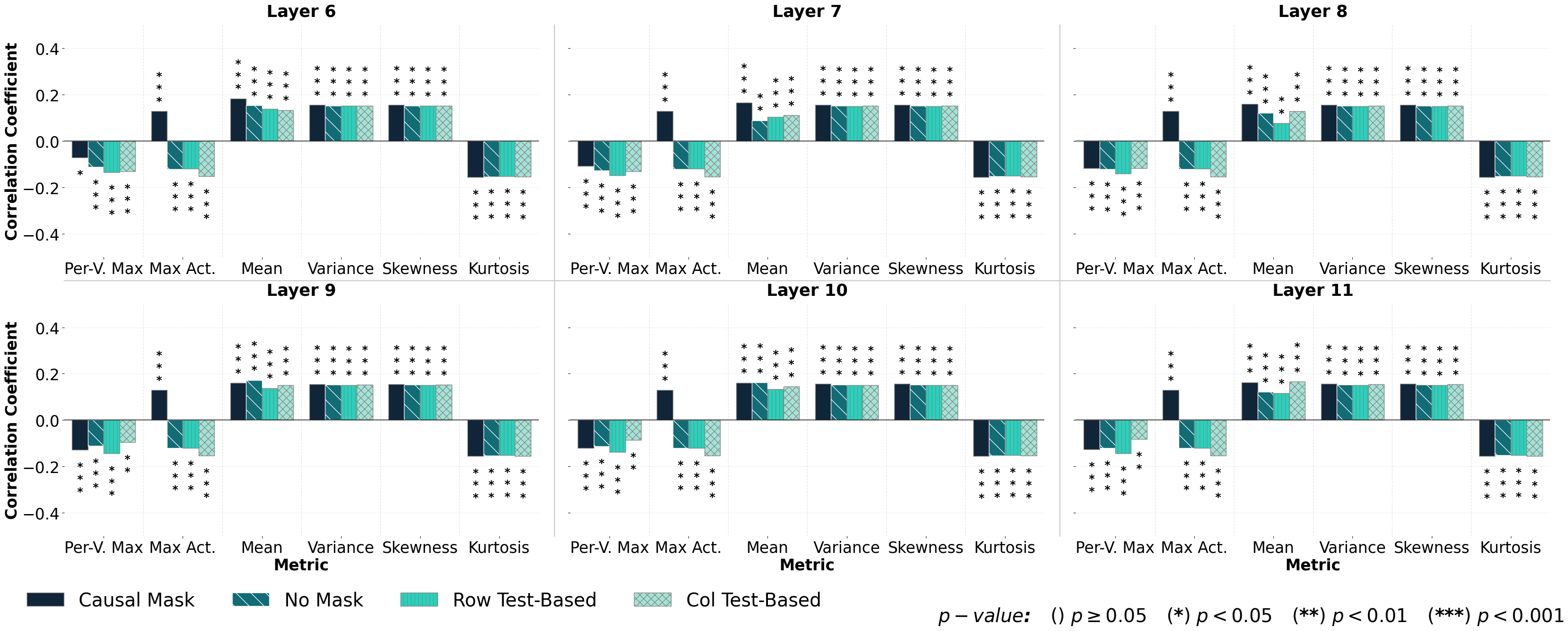}
  \includegraphics[width=\linewidth]{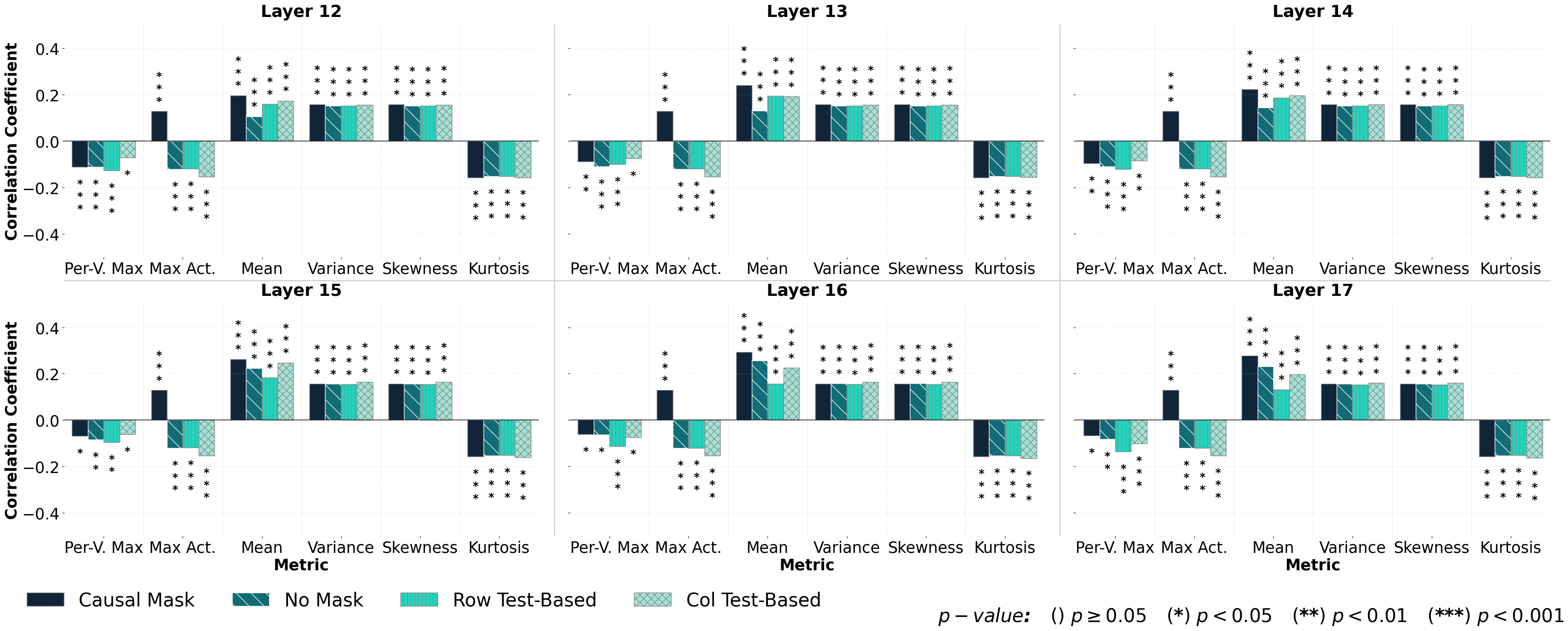}
  \caption {Shows the Spearman correlation and the associated $p$-values when computing the $pt\_score$ per-layer for embedding layer and layers 1 to 17. Results are only for Llama-3.2-3B on ARC-C.}
    \label{fig:perlayer1}
\end{figure}

\begin{figure}[t]
  \includegraphics[width=\linewidth]{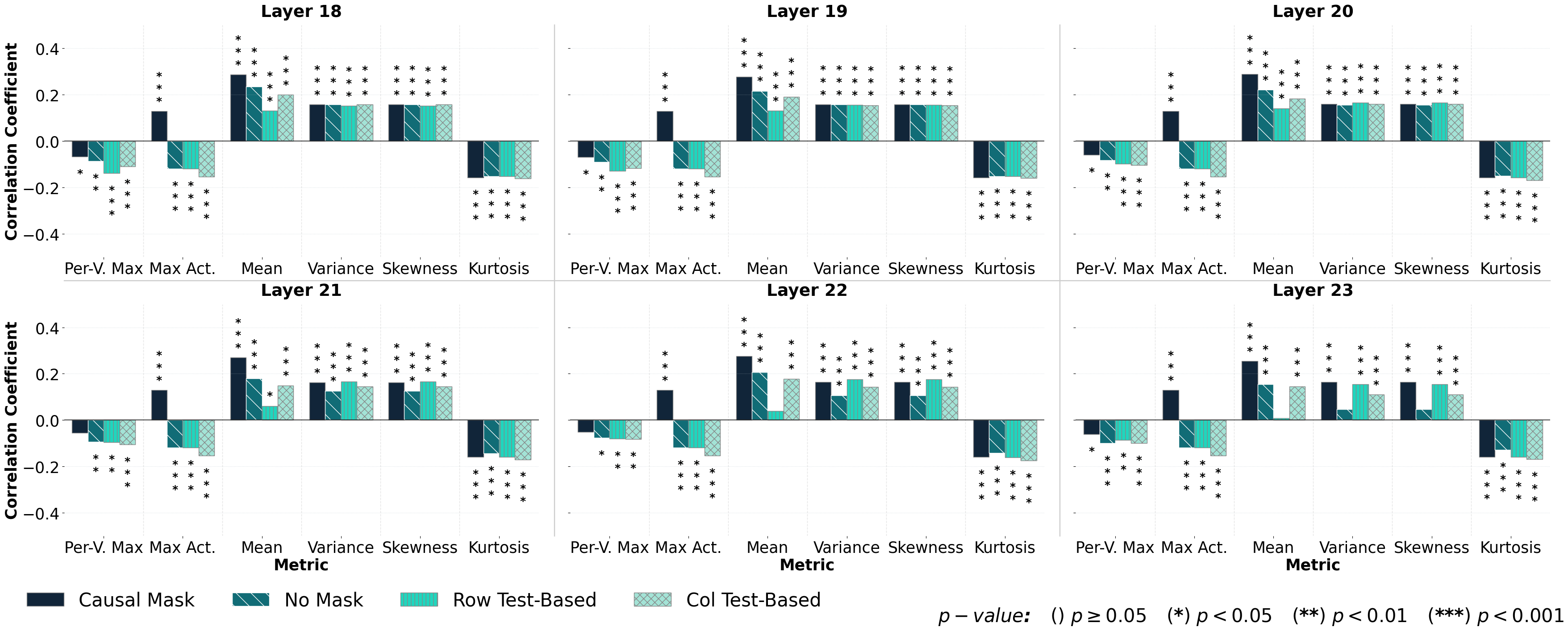}
  \includegraphics[width=\linewidth]{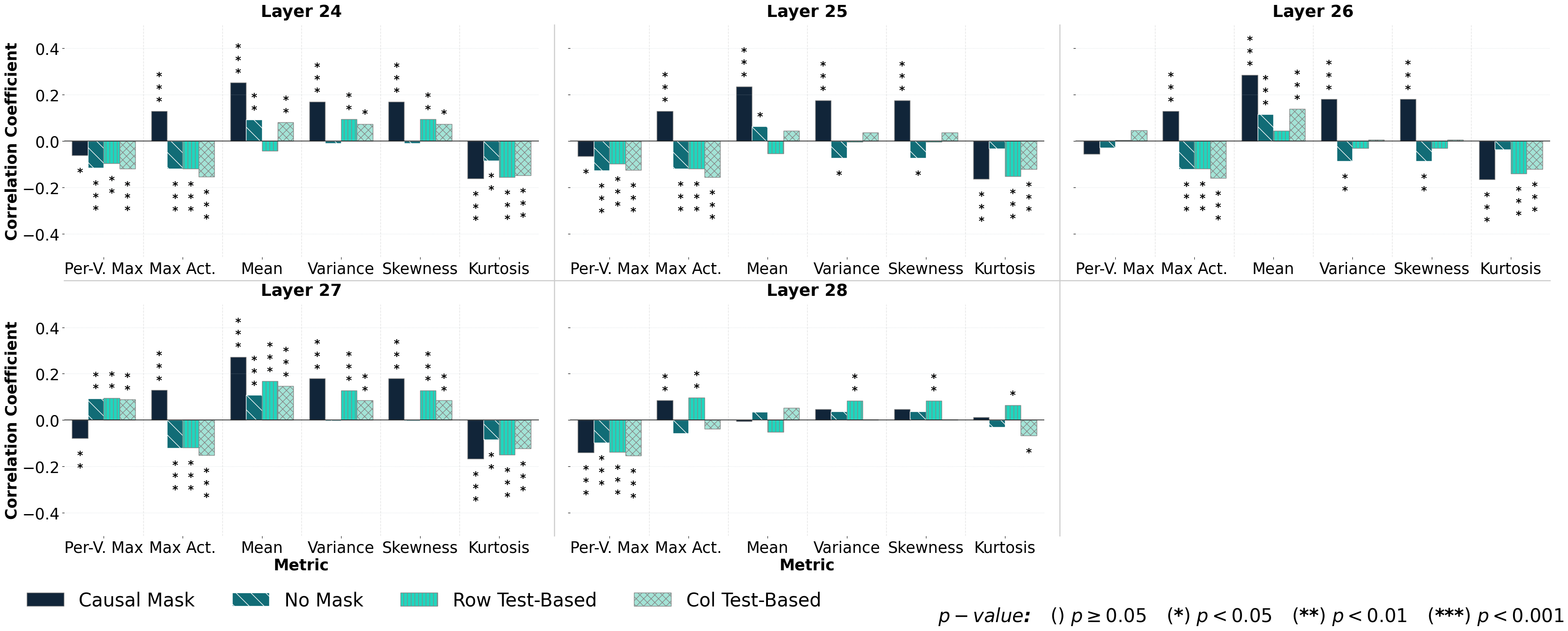}
  \caption {Shows the Spearman correlation and the associated $p$-values when computing the $pt\_score$ per-layer for layers 18 to 28. Results are only for Llama-3.2-3B on ARC-C.}
    \label{fig:perlayer2}
\end{figure}

We further calculate the scores in Equation \ref{Eq:PromptScore} for each layer independently instead of aggregating all the scores of the layers. As can be seen in Figures \ref{fig:perlayer1} and \ref{fig:perlayer2}, the lack of correlation is still present. These results are for only Llama-3.2-3B when using ARC-C as the dataset due to the high number of coefficients needed.

\end{document}